# Exploring Intensity Invariance in Deep Neural Networks for Brain Image Registration

Hassan Mahmood
*School of Science*
*Edith Cowan University (ECU)*
Joondalup, Australia
hmahmood@our.ecu.edu.au

Asim Iqbal
*Center for Intelligent Systems, EPFL*
*Neuroscience Center Zurich*
*(ZNZ), UZH/ETH Zurich*
Zurich, Switzerland
asim.iqbal@epfl.ch

Syed Mohammed Shamsul Islam
*School of Science*
*Edith Cowan University (ECU)*
Joondalup, Australia
syed.islam@ecu.edu.au

*Abstract*— Image registration is a widely-used technique in analysing large scale datasets that are captured through various imaging modalities and techniques in biomedical imaging such as MRI, X-Rays, etc. These datasets are typically collected from various sites and under different imaging protocols using a variety of scanners. Such heterogeneity in the data collection process causes inhomogeneity or variation in intensity (brightness) and noise distribution. These variations play a detrimental role in the performance of image registration, segmentation and detection algorithms. Classical image registration methods are computationally expensive but are able to handle these artifacts relatively better. However, deep learning-based techniques are shown to be computationally efficient for automated brain registration but are sensitive to the intensity variations. In this study, we investigate the effect of variation in intensity distribution among input image pairs for deep learning-based image registration methods. We find a performance degradation of these models when brain image pairs with different intensity distribution are presented even with similar structures. To overcome this limitation, we incorporate a structural similarity-based loss function in a deep neural network and test its performance on the validation split separated before training as well as on a completely unseen new dataset. We report that the deep learning models trained with structure similarity-based loss seems to perform better for both datasets. This investigation highlights a possible performance limiting factor in deep learning-based registration models and suggests a potential solution to incorporate the intensity distribution variation in the input image pairs. Our code and models are available at https://github.com/hassaanmahmood/DeepIntense.

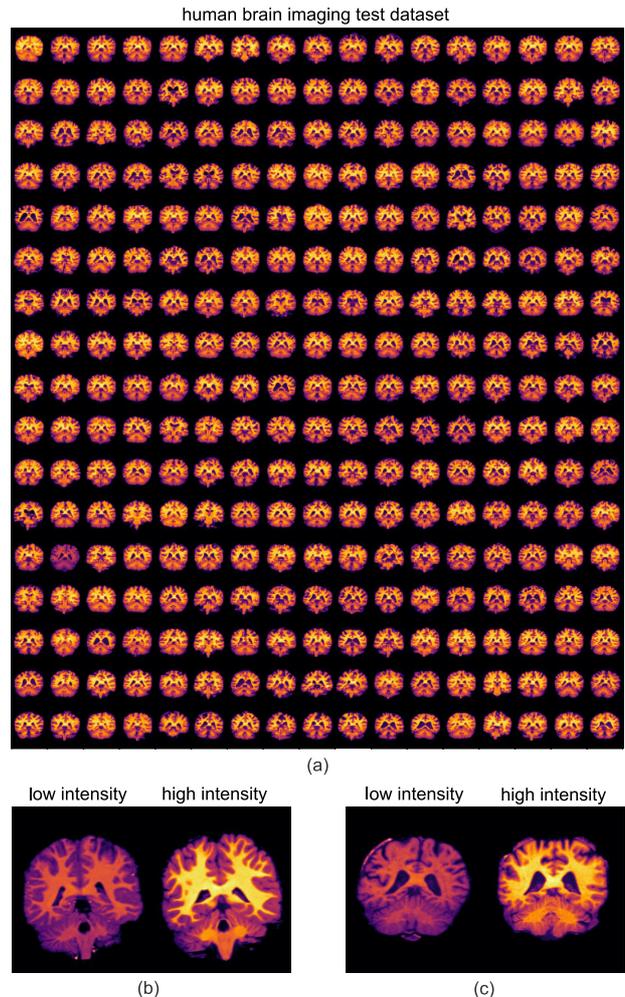

Fig. 1. Testing set of OASIS-3 brain imaging dataset. (a) 290 samples of testing dataset are shown with a large variation in intensity among them. As an example, two zoomed-in snapshots are taken as in (b) and (c) that shows a low and high intensity variance between two similar image pairs from two coronal human brain sections, captured through MRI modality.

## I. INTRODUCTION

Understanding the anatomical and functional role of different brain regions is one of the most important neuroscientific questions [1, 2, 3]. To explore this, several experiments are designed using animal models to capture the anatomical and functional correlates in different brain regions [4, 5]. In order to ensure safety, human brain imaging datasets are captured through non-invasive techniques [6] such as Magnetic Resonance Imaging (MRI) [7], X-Ray [8] and Positron Emission Tomography (PET) [9] scans that helps in exploring the anatomical and functional changes [5] in the brain under a variety of input stimuli [10]. These imaging datasets are helpful for neuroscientists to discover the difference between healthy and diseased brains [11, 12]. Analysing these brain images in a high-throughput manner is a challenging task which requires a computational method that can handle the structural and intensity-distribution variations in the brain imaging datasets [13]. As an example, Fig. 1 (a) shows the variation in the validation set extracted for our experiments from OASIS-3 [14], a commonly used publicly available human brain dataset. These images are corrected for intensity non-uniformity in the MRI data using the non-parametric non-uniform intensity normalization (N3) through the FreeSurfer [15] software. Even after normalization, we can observe the intensity difference between the image pairs as illustrated in Fig. 1 (b-c).

There has been a lot of work in the past to solve registration problem through non-learning-based image registration algorithms, [16, 17, 18, 19] just to mention a few. These algorithms typically receive a brain section image as input along with a corresponding reference image for registration that could be a brain atlas image for instance.

After receiving the input image pair, a cost function is minimized that can potentially result in an optimal transformation map for an accurate image registration with respect to the reference image. These algorithms typically perform two major steps while registering a brain image: rigid and non-rigid registrations [20]. Rigid transformation solves the global deformations such as translation, rotation, scale and sheer in input image pair whereas the non-rigid registration handles the local nonlinear deformations. Usually, rigid registration is always performed prior to the non-rigid one. These techniques have been widely used for biomedical image registration applications [21] in general and brain imaging applications in particular [22]. However, one of the major limiting factors in utilizing these techniques is their computational cost as they use iterative optimization algorithms [23].

In contrast, deep learning-based image registration methods [24] learn a function over a certain data distribution. These models are computationally efficient with a competitive performance to their classical counterparts. In recent years, deep learning models are used extensively as computational tools to understand brain's functionality in primates [25, 26] and rodents [27, 28, 29] as well as decoding the functionality of brain signals [30, 31, 32, 33]. While these methods are also performing state-of-the-art in brain image registration [34] and segmentation [35] on diverse datasets yet they lack a generalisability even within the same data distribution which is used for training. The general principle for working of these deep learning-based registration models is to find the displacement field as in the optical flow. It is also clear from the optical flow perspective that if there exists a brightness change in between the two consecutive frames, the computed flow is likely to be erroneous [36] because the optical flow algorithms work under an assumption that the corresponding pixel values in two consecutive frames remain constant.

These limitations in deep learning-based registration frameworks can be addressed in several ways: one of the ways to make these models more generalised is to add more training or augmented data but in practice there is always a chance to miss a certain data distribution. The other way is to come up with the new deep learning-based architectures and innovative loss functions. We make an effort to contribute towards the latter by introducing a structural similarity-based [37] loss function during the training of the network. Predominantly, structural similarity index measure (SSIM) is used in the perceptual quality measurement of images.

To the best of our knowledge, this study is a pioneering effort to rigorously investigate the relationship between spatial intensity distribution of input pairs and its effects on deep learning-based registration networks as well as introducing structural similarity metric in a loss function for deep learning-based image registration techniques. We hypothesizes that only intensity-based loss will be insufficient for the network's layers to learn the intensity invariant features during training. With an addition of the structural loss term, our results show that deep learning-based models converge better to handle the difference in intensity distributions of input pair images.

After introduction, the structure of the paper is laid as follows: we start with a background followed by the methods,

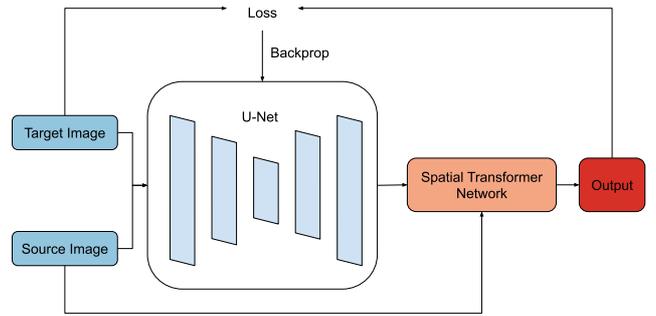

Fig. 2. Deep learning-based image registration network. An input source and target images are fed to the U-Net, followed by a Spatial Transformer Network that results in the registered output image

experiments and results. We conclude after a discussion and towards possible future research directions.

## II. BACKGROUND

### A. Image registration

The purpose of image registration is to compute a spatial transformation or displacement field ($\Upsilon$) by which a source image ($I_s$) can be moved to a target image ($I_t$). In traditional image registration, this task is achieved through an iterative process by optimising over a certain similarity metric ($S$). The main goal is to minimise (1) each time for every new image pair to find an optimal displacement field ($\Upsilon$), here, $\Upsilon$ can be used to warp the $I_s$ or corresponding segmentation and can potentially be used for any further analysis.

$$|I_t - (I_s \cdot \Upsilon)| \tag{1}$$

Finding displacement fields and calculating optical flow are similar in nature [38, 39]. With the introduction of deep learning-based optical flow methods, researchers have started to explore similar network architectures for image registration in the computer vision field in general and medical imaging domain in particular. Early learning-based networks were mostly supervised [40, 41] i.e. these models needed a ground truth displacement field. With the introduction of spatial transformer network (STN) [42], researchers have also started to explore the unsupervised deep learning-based registration architectures [43, 44, 45].

### B. Deep learning-based image registration

In deep learning-based registration models, the goal is to estimate the displacement field by computing a function over ($I_t$) and ($I_s$) in a generally single forward computation. To achieve this, a deep network is trained over a set of images ($I$), where $I \subset R^3$ or $I \subset R^2$. A function, $G_\vartheta(I_t, I_s) = \gamma$ is learned through encoder/decoder or U-Net [46] type architecture, where $\Upsilon$ is a displacement field which is used to warp $I_s$. During training of $G_\vartheta$, such parameters $\vartheta$ of deep networks are to be searched which will minimise the loss function ($L$) based on some similarity metric. This optimisation can be described as follows:

$$\hat{\gamma} = \underset{\gamma}{\mathrm{argmin}}\{L(I_t, \bar{I}_s) + R(\gamma)\} \tag{2}$$

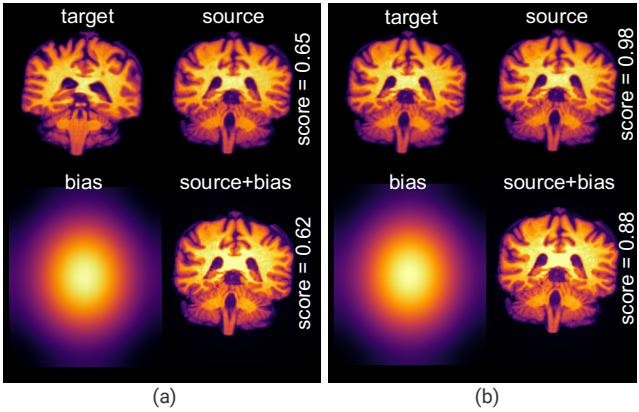

Fig. 3. Effect of intensity variation in image registration. (a) Registration of target image without and with adding illumination bias to the source image. (b) Registration of target image without and with adding illumination bias to the source image (here source image is the same as target image).
Note: All the brain images are scaled to the same data range for display purpose with the exception of the Gaussian disc images, bottom left in Fig. 3 (a-b) otherwise these Gaussian discs will not be visible to the naked eye.

where, $\bar{I}_s = \gamma \cdot I_s$ and $R$ is any kind of regularisation on the displacement field.

Deep learning-based registration models learn the transformation function beforehand for a certain image data distribution. During the testing time, these learned models can predict the transformations. The output of these architectures is the displacement field which is further fed into the spatial transformation layer to warp the moving image. A high-level overview of deep learning-based image registration model is visually represented in Fig. 2.

### C. Effect of variation in intensity distribution

As these deep learning networks are inspired by optical flow-based architectures in general, they acquire an inherent limitation to perform adequately in case of illumination distribution variance for a given pair of input images. We also find in our initial experiments that a minor change in illumination (brightness) of MRI scan impart a detrimental effect on the image registration accuracy, observed through a decreased DICE score. We generated an illumination bias by convolving a point image with a Gaussian circular disc as shown in the bottom left of Fig. 3 (a-b). We conducted two kinds of experiments to validate our hypothesis. In experiment#1 (Fig. 3 (a)), we took two different source and target images and computed the DICE score before and after adding illumination bias to the source image by keeping the target image the same. We observe a ~5% drop in DICE score (from 0.65 to 0.62). The source image (post-bias) can be seen in the bottom right of Fig. 3 (a).

In experiment#2, we did the same as in experiment#1 but this time we kept the target and source images the same. We observed a performance drop in DICE score by ~10% (from 0.98 to 0.88). These findings clearly show the effect of difference of intensity (brightness) distribution and a need to conduct further investigation on deep learning-based registration networks.

## III. METHODS

Our main focus is to investigate the behaviour of deep learning-based image registration models, given a difference of intensity distribution in the input image pair. For this purpose, we took advantage of an implementation of VoxelMorph [34] in PyTorch [47]. The architecture for our experimental models is based on U-Net. We are working with images ($I \subset R^2$), so our input is of the size 160x192x2. The network is trained using Adam optimization technique [48] with learning rate set to 0.0001 with a batch size of 5. All models were trained on NVIDIA's GeForce GTX 1080 Ti GPU with a stopping criteria of delta=0.0000001 and patience=25.

We trained all the models for subject to subject scenario as compared to subject to atlas. This is one of the core problems in medical image registration as pointed in [34]. To achieve this, all input pairs are randomly chosen for the batches during each iteration.

### A. Loss function

The convergence of image registration algorithms depends upon a similarity metric as it acts as an objective function to estimate the quality of registration by computing the similarity between the target image and the source image. We analysed various combinations of different similarity metrics in the loss function, whereas, each similarity metric computes certain characteristics from the image pair. In the literature, the most widely used similarity metric is the mean squared error (MSE) and the cross-correlation (CC). MSE is computed for $I_t$ (target image) and $\bar{I}_s$ (source image) as follows:

$$MSE(I_t, \bar{I}_s) = \frac{1}{n}\sum_{i=1}^{n}\left(I_t(i) - \bar{I}_s(i)\right)^2$$

(3)

Here, $i$ is the location of each pixel. Similarly, the CC is computed as follows:

$$CC = -\left(\frac{\sum_i [I_t[i] - \overline{I_t}][\bar{I}_s[i] - \overline{\bar{I}_s}]}{\sqrt{\sum_i (I_t[i] - \overline{I_t})^2}\sqrt{\sum_i (\bar{I}_s[i] - \overline{\bar{I}_s})^2}}\right)$$

(4)

CC is the measure of similarity and it will return a maximum value when two input images are identical. To use this as a loss function for training in a deep neural network, we multiply it with a negative sign since we want to minimize our loss. We introduce a structural similarity metric-based loss in our total loss equation for training the deep learning-based registration network. The Structural Similarity Index Measurement (SSIM) is initially introduced for image quality assessment [37] and later utilized in the traditional image registration to some extent [49]. SSIM is a combination of three different kinds of measurements: brightness comparison, contrast comparison and structural comparison. Brightness comparison (*b*) is computed through the mean ($\mu$) of intensity of a given image, *I*.

$$\mu_I = \frac{1}{n}\sum_{i=1}^{n} I_i \tag{5}$$

$$b(I_t, \bar{I}_s) = \frac{2\mu_{I_t}\mu_{\bar{I}_s} + C1}{\mu_{I_t}^2 + \mu_{\bar{I}_s}^2 + C1} \tag{6}$$

The Contrast comparison (*c*) uses the standard deviation ($\sigma$).

$$\sigma = \sqrt{Var(I)} \tag{7}$$

$$c(I_t, \bar{I}_s) = \frac{2\sigma_{I_t}\sigma_{\bar{I}_s} + C2}{\sigma_{I_t}^2 + \sigma_{\bar{I}_s}^2 + C2} \tag{8}$$

And the third component, the structural comparison (*st*) is computed based on the covariance of $I_t$ and $\bar{I}_s$ as follows:

$$st(I_t, \bar{I}_s) = \frac{\sigma_{I_t \bar{I}_s} + C3}{\sigma_{I_t}\sigma_{\bar{I}_s} + C3} \tag{9}$$

Here, $C1 = (0.01)^2, C2 = (0.03)^2, C3 = C2/2$

These statistics are calculated within a local window. We experimented with different window sizes (5, 7, 11) and did not observe much effect in the performance so we randomly kept the window size of 11 for all the experiments.

Now, the final SSIM is:

$$SSIM(I_t, \bar{I}_s) = b(I_t, \bar{I}_s) \cdot c(I_t, \bar{I}_s) \cdot st(I_t, \bar{I}_s) \tag{10}$$

Similar to CC, SSIM is a similarity metric, and it will return maximum output on two similar images. To use this in a loss function, we will use: $1 - SSIM$. In SSIM, the three components are relatively independent [37] that is the main reason for testing the performance of deep neural networks trained on the loss based on this metric. As from its definition, these three parts are relatively independent of each other and can guide the network during training towards supposedly a better solution against the difference of brightness distribution in the input image pair. Just to elaborate further about the effect of selecting different similarity metrics, Fig. 4 shows a comparison of mean intensity difference after registration with a baseline model. All scores are sorted by the non-rigid (L1) score in an ascending order.

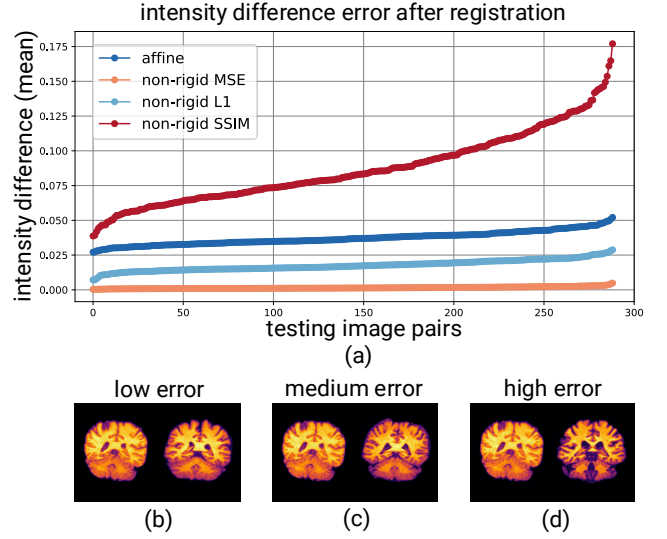

Fig. 4. Comparison of mean intensity difference error after registration among different similarity metrics. (a) A quantitative comparison of a randomly selected target image with the entire testing set using different similarity metrics after affine and non-rigid registration, (b, c, d) shows three corresponding sample pairs from left, middle and right part of the plots.

Furthermore, to avoid the unrealistic predicted deformation, a regularization term (*R*) is added in the loss, which is based on the gradient of the displacement field.

The total loss equation with MSE for our experimentation can be described as below:

$$Loss\_MSE\_SSIM = \alpha\big(1 - SSIM(I_t, \bar{I}_s)\big) + (1-\alpha)MSE\big(I_t, \bar{I}_s\big) + \beta R \tag{11}$$

The total loss equation with CC is defined as:

$$Loss\_CC\_SSIM = \alpha\big(1 - SSIM(I_t, \bar{I}_s)\big) + (1-\alpha)CC\big(I_t, \bar{I}_s\big) + \beta R \tag{12}$$

Here, $\alpha$ is a loss ratio factor, $\alpha=[0,1]$ and $\beta$ is to enforce the regularization. For our experiments, we kept $\beta=0.01$.

*B. Brain image data collection and preparation*

In this work, we use publicly available T1 MRI data from OASIS-3 [14] and ABIDE [50] datasets. The data is already resampled to a 256×256×256 grid with 1mm isotropic voxels. OASIS-3 and ABIDE datasets are also provided with different segmentations for both the left and right hemisphere of the brain.

Our current analysis is for 2D brain slices so we take every 100[th] coronal slice from each brain in OASIS-3 and ABIDE datasets. After cropping the resulting image to 160×192, we perform the image normalization and affine registration to a fixed 2D reference. We randomly select 1656 slices for training from OASIS-3 2D slices and 290 slices for testing purposes from both OASIS-3 and ABIDE datasets.

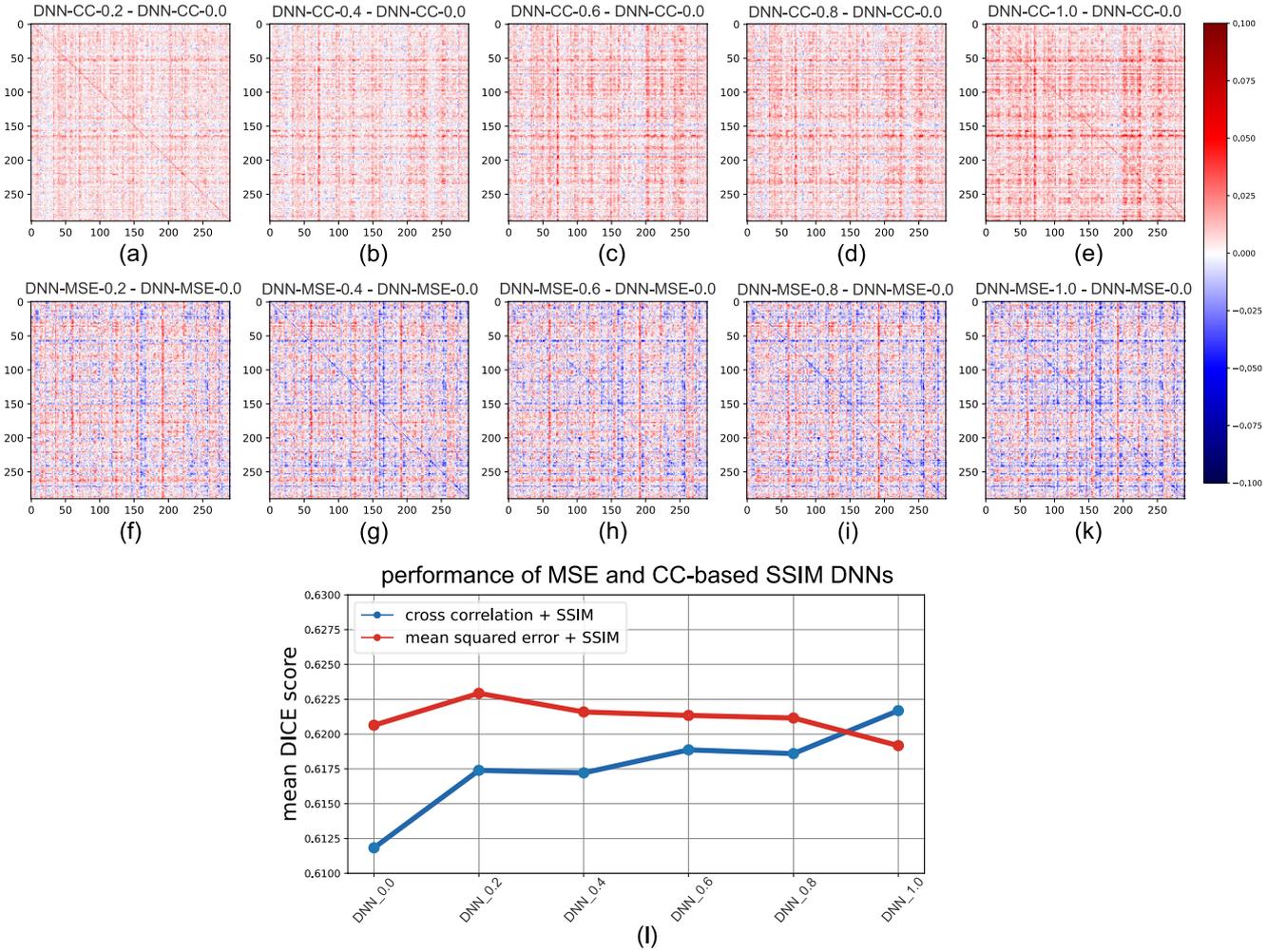

Fig. 5. Performance comparison between the baseline model (DNN_0.0: VoxelMorph (MSE, CC)) and our structural similarity-based deep learning models. (a-e) A quantitative performance of $\Delta_{\text{Dice\_cc}}$ for each test image pair by introducing different loss ratios ($\alpha$) in (12). (f-k) A performance comparison of $\Delta_{\text{Dice\_mse}}$ by adding different loss ratios ($\alpha$) in (11). (l) Performance of the models are shown as measured through mean DICE scores on the entire testing dataset of OASIS-3.
Note: Herein, DNN_0.0, 0.0 corresponds to the loss ratio ($\alpha$) in (11-12).

## IV. RESULTS AND DISCUSSION

As our focus is on the subject to subject image registration problem so our experiments are designed in a similar manner. To deeply analyse the trends, we perform the registration for each image with all the images in a testing set of 290 samples. In this way, we are getting 290 different target images and we calculated their respective DICE scores. This will result in a 290x290 matrix in which each row contains the registration result of the target image with the 290 testing (source) images. Let $I_t$ be the target image from the first index of testing data i.e. $ind_{I_t} = 0$ then $ind_{I_s} = [0, 289]$ will be the locations of all source images to be registered and the same will be repeated with $ind_{I_t}$ reaching until 289. We calculate this 290x290 matrix for each model where the baseline model is VoxelMorph with MSE and CC losses. Afterwards, we calculate the difference between our structural similarity-based networks' DICE scores with the baseline models which again results in a 290x290 matrix:

$$\gamma_{voxel\_mse} = G_{\vartheta_{voxel\_mse}}(I_t, I_s) \quad (13)$$

$$Dice_{voxel\_mse} = Dice(B_t, \gamma_{voxel\_mse} \cdot B_s) \quad (14)$$

$$\gamma_{voxel\_cc} = G_{\vartheta_{voxel\_cc}}(I_t, I_s) \quad (15)$$

$$Dice_{voxel\_cc} = Dice(B_t, \gamma_{voxel\_cc} \cdot B_s) \quad (16)$$

$$\Delta_{\text{Dice\_mse}} = Dice_{SSIM\_mse} - Dice_{voxel\_mse} \quad (17)$$

$$\Delta_{\text{Dice\_cc}} = Dice_{SSIM\_cc} - Dice_{voxel\_cc} \quad (18)$$

Here, $B$ is the corresponding segmentations of the input pair. We compute $\Delta_{\text{Dice\_mse}}$ and $\Delta_{\text{Dice\_cc}}$ for all possible test input pair combinations. This gives us effectively a matrix as shown in Fig. 5 and we can clearly see the justified outcome at the diagonal of each matrix i.e. we are getting a consistent trend as both $I_t$ and $I_s$ are the same images. Furthermore, we can observe the trend in the $\Delta_{Dice\_cc}$ as we increase the weight of structural similarity in the loss function (12). Fig. 5 (a-e) shows an incremental trend which is also shown (as a mean of the whole 290x290 DICE matrix for each CC+SSIM-based model separately) in Fig. 5(l) as a blue plot. A somewhat opposite trend is observed in Fig. (f-k) where $\Delta_{Dice\_mse}$ is shown as we change the weight of the structural similarity-based loss function (11). Similarly, the mean of DICE metric (MSE+SSIM) is shown in red in the Fig. 5 (l). Although, the trend of the models trained with MSE and CC-based SSIM loss functions seem diverging as we increase the $\alpha$ terms in our loss function but in both cases the performance is improved as compared to the baseline model. We observe a similar trend in ABIDE dataset as shown in Fig. 7 where we observe an increment in the performance of SSIM-loss-based model on the previously unseen dataset. However, this requires further investigation to model the effect of $\alpha$ in different loss functions as well as for other imaging datasets.

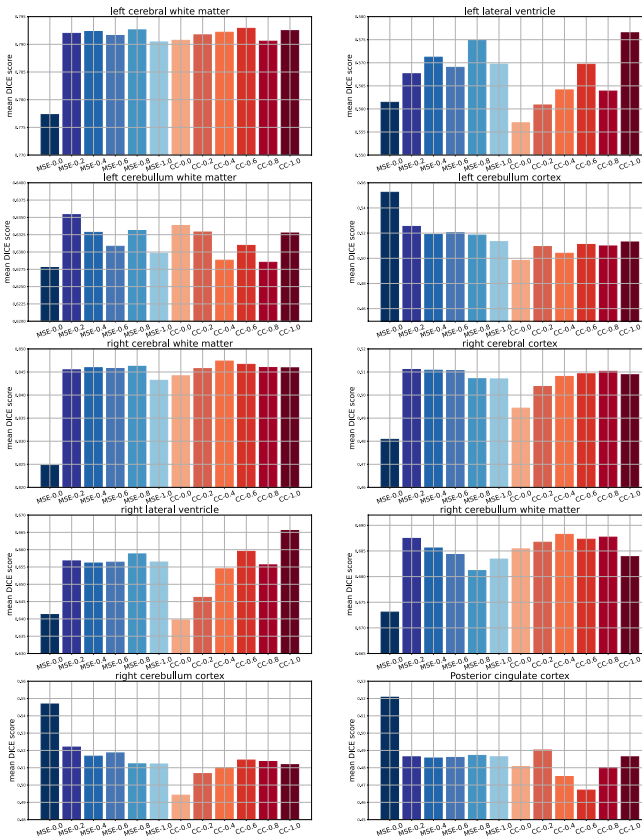

Fig. 6. Performance of MSE and CC-based models with SSIM loss function. Average DICE score of 10 brain regions are plotted with their names on top of each panel. Except three regions (left cerebellum cortex, right cerebellum cortex and posterior cingulate cortex), SSIM-based deep learning models are shown to perform better than the baselines models i.e. VoxelMorph.

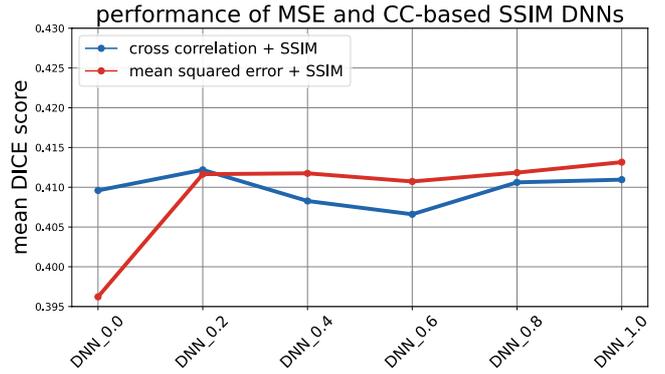

Fig. 7. Performance of the models are shown as measured through mean DICE scores on the entire testing dataset of ABIDE.

We further investigate the performance of all models in segmenting individual brain regions. Henceforth, we select the brain regions that are common in all the testing samples. Fig. 6 shows the bar plots of all the models for 10 different brain regions such as left cerebral white matter, right lateral ventricle, etc. in the OASIS-3 dataset. Adding structural similarity-based loss function in the deep neural networks results in better performance in 7 out of 10 brain regions in the entire testing set. This shows that in cases where a traditional deep learning-based registration model is unable to capture the structurally relevant features at varying intensity, model trained with SSIM-based loss function is robust to capture the intensity invariance in diverse brain regions.

## V. CONCLUSION AND FUTURE WORK

In this study, we highlight the limitation of deep learning-based registration methods in tolerating the difference of intensity distribution in the pair of input images. We propose a simple solution that can add intensity invariance in these models by introducing a structural similarity-based loss function. We test the performance of SSIM-based loss function on a variety of brain imaging samples. The results show a promising effect of introducing structural similarity-based loss functions in deep learning-based models that can handle the intensity invariance in brain imaging datasets. Although, we observe an increment in the overall performance of the models but the results can further be improved by exploring more innovative loss functions that can simultaneously handle different invariancies in the imaging datasets. In future, we aim to investigate it further by experimenting with different loss functions as well as observing their performance on a variety of brain imaging datasets of humans as well as mouse. Furthermore, we aim to extend this work to other deep learning-based network architectures to investigate the similar effect due to intensity distribution variation.